%% file: main_workshop.tex
\documentclass{article}

\PassOptionsToPackage{numbers, compress}{natbib}

\usepackage[final]{neurips_distshift_2022}

\usepackage[utf8]{inputenc} 
\usepackage[T1]{fontenc}    
\usepackage{hyperref}       
\usepackage{url}            
\usepackage{booktabs}       
\usepackage{amsfonts}       
\usepackage{nicefrac}       
\usepackage{microtype}      
\usepackage{xcolor}         

\usepackage{graphicx}

\usepackage{booktabs}
\usepackage{multirow}
\usepackage{algorithm}
\usepackage{algpseudocode}
\usepackage{comment}
\input{math_commands}

\newcommand{\eg}{\textit{e}.\textit{g}. }

\definecolor{cyan}{rgb}{0,1,1}
\definecolor{tableex}{rgb}{0,0,0}
\definecolor{algo_comments}{HTML}{297929}
\newcommand{\algrule}[1][.2pt]{\par\vskip.3\baselineskip\hrule height #1\par\vskip.3\baselineskip}
\newcommand*\samethanks[1][\value{footnote}]{\footnotemark[#1]}

\newcommand*{\tabindent}{\hspace*{0.5cm}}%

\title{Env-Aware Anomaly Detection: Ignore Style Changes, Stay True to Content!}

\author{%
  Stefan Smeu\thanks{Equal contribution} \\
  Bitdefender, Romania\\
  University of Bucharest\\
  \texttt{ssmeu@bitdefender.com} \\
  \And
  Elena Burceanu\samethanks \\
  Bitdefender, Romania \\
  \texttt{eburceanu@bitdefender.com} \\
  \And
  Andrei Liviu Nicolicioiu \\
  MPI for Intelligent Systems, T\"ubingen \\
  \texttt{andrei.nicolicioiu@tuebingen.mpg.de} \\
  \And
  Emanuela Haller\\
  Bitdefender, Romania \\
  \texttt{ehaller@bitdefender.com} \\
}

\begin{document}

\maketitle

\begin{abstract}
We introduce a formalization and benchmark for the unsupervised anomaly detection task in the distribution-shift scenario. Our work builds upon the iWildCam dataset, and, to the best of our knowledge, we are the first to propose such an approach for visual data. We empirically validate that environment-aware methods perform better in such cases when compared with the basic Empirical Risk Minimization (ERM). We next propose an extension for generating positive samples for contrastive methods that considers the environment labels when training, improving the ERM baseline score by 8.7\%.

\end{abstract}

\section{Introduction and related work}

Identifying and following novelty \cite{mather2013novelty} is an intriguing human ability that could trigger scientific discoveries \cite{kuhn1970structure}. Machine learning models that can mimic this behavior and detect novelty when facing unfamiliar situations are vital for fields like video surveillance~\cite{ionescu}, intrusion detection in cybersecurity~\cite{kwon2019survey}, manufacturing inspection~\cite{lopez2017categorization}, and many others~\cite{ruff2021unifying_ad_survey}. 
Anomaly Detection (\textbf{AD}) is an umbrella term \cite{ood-survey} for methods whose goal is to identify samples that deviate from an assumed notion of normality. Normals and anomalies are supposed to come from different distributions. But how and up to what limit do they differ? Defining what changes constitute anomalies and what changes should be ignored is essential. 

Deep learning methods proved their representation power in multiple fields \cite{derma_dl, cardio_dl, dalle2, bert, resnet, vit, dino, cybersec_dl_survey} and were assumed to become invariant to irrelevant aspects under the big data regime. Yet, recent works proved that these representations are susceptible to unwanted biases~\cite{beery2018recognition} and prone to finding shortcuts~\cite{geirhos2020shortcut}, relying on spurious features while failing to capture relevant aspects of the data. Consequently, those models exhibit poor performance when dealing with slightly different, out-of-distribution (OOD) settings, where spurious features are no longer informative. Avoiding spurious correlations is a challenging problem, impossible to solve in the in-distribution (ID) training setup \cite{scholkopf2021toward}. Recent works \cite{ muandet2013domain, peters2016causal, wilds, mind_the_gap, clear, anoshift, irm, lisa} tackle this problem by using an informative process that splits the dataset into multiple environments, extracting additional information. Except for AnoShift~\cite{anoshift}, which focuses on network traffic data, all those benchmarks and approaches address supervised scenarios.

We follow the same line to enable robust AD and exploit the multi-environments approach in unsupervised anomaly detection from visual data. Moreover, we formalize the notion of anomaly under the distribution shift paradigm.

To summarize, our \textbf{main contributions} are the following:
\begin{itemize}
    \item We propose a benchmark for unsupervised anomaly detection in images, focusing on real-world cases, where the input distribution is different in sub-groups of data. We formally emphasize the differences between anomaly detection and classical (supervised) distribution shift analysis.
    \item We validate that shallow AD methods can benefit from working on top of embeddings pretrained using environment-aware methods (like Fish, IRM, LISA). We prove consistent improvements over ERM pretraining over a wide range of AD methods.
    \item We introduce a way of adjusting the contrastive methods to be aware of multiple environments, making them more robust in out-of-distribution setups. Empirical validation over MoCo v3, shows an 8.7\% increase in ROC-AUC w.r.t. ERM, on iWildsCam dataset, in the anomaly detection setup. 
\end{itemize}

\begin{figure}[t!]
\begin{center}
\centering
    \includegraphics[width=0.95\textwidth]{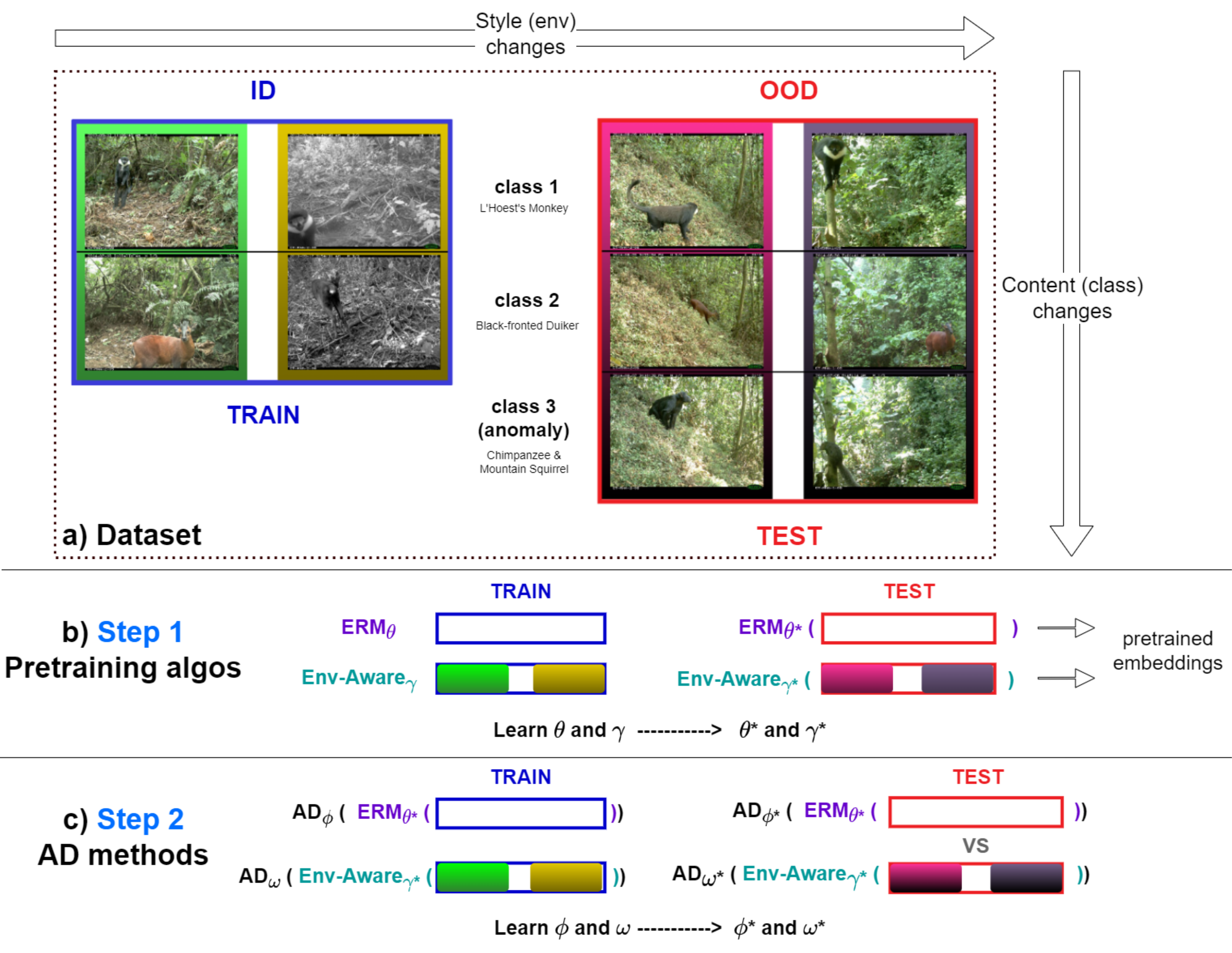}
\end{center}
\caption{OOD Content and Style Setup. a) Dataset: The samples' input distribution varies on the Content and Style axis. In training, we have only normal data, while in the testset we also have a 3rd (anomalous) class. b) Step 1: Pretraining algos learn its parameters using the training data, with labels for the two normal classes. c) Step 2: AD methods use the embeddings learned in Step 1 to transform their input.}
\label{fig:setup}
\end{figure}

\section{Generalization facets for Anomaly Detection}



\textbf{Latent factorization of the data } It is useful to formalize the data samples $x$ as being determined by two latent factors: Content and Style~\cite{mitrovic2021representation_relic_v1, julius2021self_supp}. The \textbf{Content} should determine the task at hand, i.e., it should be the cause of the desired target. At the same time, the \textbf{Style} could represent unrelated features spuriously correlated with the target.
Inferring these latent factors is an extremely difficult problem, seen as a goal of representation learning \cite{mitrovic2021representation_relic_v1}. It is impossible in the unsupervised setting without additional inductive biases, or other information \cite{hyvarinen1999nonlinear, locatello2019challenging} and it is outside our scope. Instead, we start from a weaker assumption, that we have data in which only the \textbf{Style} is changed. We aim to use this factorization in AD to highlight directions toward building methods that are robust to irrelevant changes (involving Style) while capable of detecting relevant changes (involving Content).

\textbf{Environments } We call \textit{domains} or \textit{environments} \cite{wilds, irm} sub-groups of the data, each with a different distribution, but all respecting some common basic rules. Namely, the Content is shared, but Style or relations involving Style change. Examples of domains include pictures taken indoor vs. outdoor \cite{ade20k},  or in different locations \cite{iwildcam}, real photos vs sketches \cite{pacs}, or images of animals with changing associations in each environments~\cite{metashift}. Our goal is to be robust to the Style differences between different environments while identifying the Content changes as anomalies.

\subsection{Out-of-distribution regimes } When dealing with real-world data, the test distributions usually differ from the training ones, encountering changes in Style or/and Content. We provide next an in-depth characterization of possible scenarios for AD in those regimes, linking them to common methods that work in each category for supervised tasks. For explicit examples and details, see Appendix~\ref{apx:formalize-sup-unsup}.

\textbf{A. ID setting } The default paradigm in Machine Learning, both in supervised and unsupervised learning. 
Although this is the default paradigm, the usual assumption that train and test data come from the same distribution is very strong and almost never true for real-world datasets \cite{derma_ood, dataset_bias_torralba, anoshift, clear, wilds}. 

\textbf{B. Style OOD } Most works that develop methods robust to some (i.e. Style) distribution changes reside in this category \cite{fish, irm, rex, lisa}. Environments have differences based on Style, but have the same Content and the goal is to learn representations that are invariant across environments.

\textbf{C. Content OOD } The assumption here is that environments contain changes in distribution that are \textit{always relevant} (i.e. changes in Content) for the task and should be noticed. Methods in this category must detect such changes while optionally performing another basic task. Anomaly, novelty, or OOD detection methods work in this regime ~\cite{ood-survey}.

\textbf{D. Style and Content OOD } Here, environments bring changes in \textit{both} Content and Style. We argue that this is the most realistic setting and it is mainly unaddressed in the anomaly detection literature. An ideal anomaly detection method will only detect Content anomalies, while being robust to Style changes. Our main analyses and experiments are performed in this setting, showing the blind spots of current approaches and possible ways forward. 

We formalize and detail the distribution shifting scenarios in Appendix~\ref{apx:formalize-sup-unsup}. To the best of our knowledge, we are the first to cover this topic for anomaly detection in particular and for unsupervised learning in general.




\begin{figure}[t!]
\begin{center}
    \includegraphics[width=0.95\textwidth]{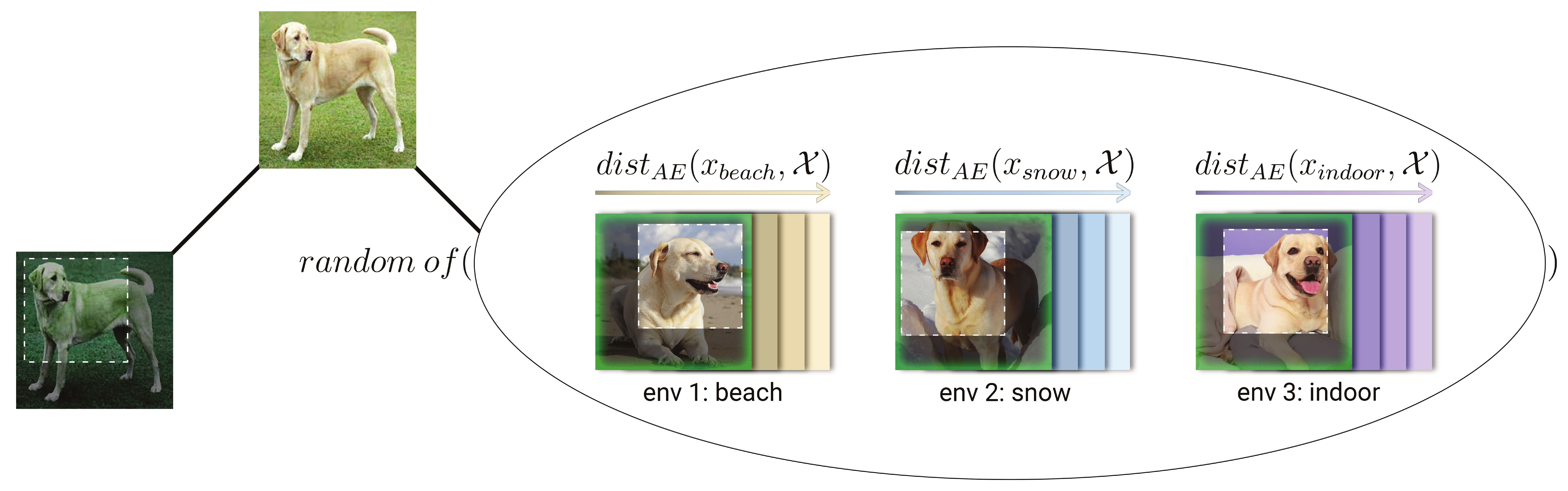}
\end{center}
\caption{Positive samples in contrastive learning. In EA-MoCo, our env-aware baseline over MoCov3, for the positive samples we use: the basic augmented anchor (left) and the closest sample from another, randomly chosen, environment (right). We compute distances over representations obtained with an diffusion based autoencoder, learned under ERM, over samples from all environments.}
\label{fig:ea-moco}
\end{figure}

\subsection{Our approach}
\label{sec:our_approach}
We are interested in detecting anomalies in the most realistic scenario, \textbf{D. Style and Content OOD}, where both the Style and the Content can change between environments, and our goal is to detect Content changes, while learning to be robust to Style changes. Generally, AD methods work on top of pretrained embeddings. We first show how to adapt envs-aware methods~\cite{irm, fish, lisa} to an unsupervised AD task (using a non-AD task for pretraining). Next, we introduce our self-supervised environment-aware method: EA-MoCo, for pretraining the AD embeddings.

\textbf{Proposed setup } We propose a pretraining approach for learning env-aware embeddings that takes advantage of the env information in some way or another. Current methods are using the environment labels in training, to build env-invariance. Similarly to them, but unlike other AD methods, we use envs in the pretraining phase to learn better representations for the downstream task (anomaly detection in our case). We empirically prove (Sec.~\ref{sec:experiments}) that learning embeddings robust to domain changes (Style) in the pretraining phase improves the overall performance of the downstream AD task. For a better understanding of the setup, we present a visual representation in Fig.~\ref{fig:setup}.

\textbf{Supervised non-AD pretraining } First, we adapt already existing env-aware solutions for supervised learning (\eg IRM~\cite{irm}) to anomaly detection. For creating the supervised task needed for these methods, one can use additional labels found in the dataset or can create pretext tasks. In particular, we divide the dataset samples into 3 sets: 2 normal classes and 1 anomalous. We use the normal ones for modeling a binary classification task. Please note that the task is supervised to predict one of the two normal classes, not directly supervised for the Anomaly Detection task. This way, we do not need or use anomaly labels. We train the env-aware method on this binary classification task to learn embeddings and apply AD methods on top of those learned embeddings.

\textbf{Fully unsupervised pretraining } To overcome the need for supervision, we propose \textbf{EA-MoCo}, an env-aware contrastive learning approach. Briefly, we train an diffusion based autoencoder~\cite{diffusion} over all training envs, and based on the learned representations, we compute distances between all samples. Next, we define the contrastive learning objective by selecting pairs of samples from different randomly sampled domains which are close to each other, considering our previously defined distances. See Fig.~\ref{fig:ea-moco} and Appendix~\ref{apx:ea-moco-env} for details.

We employ AD methods on top of our supervised and unsupervised embeddings, proving large improvements when using the env information in pretraining (Sec.~\ref{sec:experiments})

\section{Experimental results}\label{sec:experiments}

\begin{figure}[t]

\begin{center}
    \includegraphics[width=0.49\textwidth]{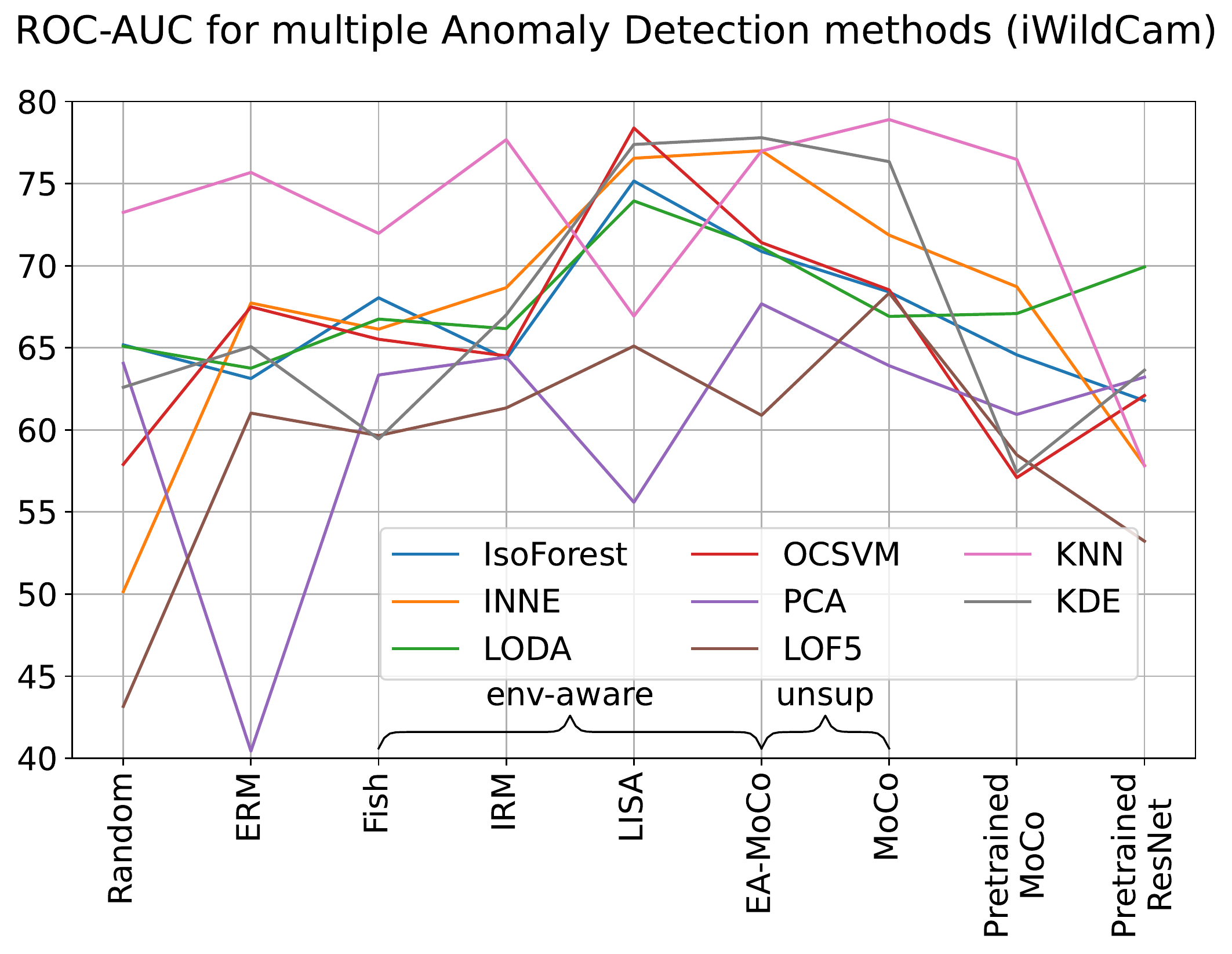}
    \includegraphics[width=0.49\textwidth]{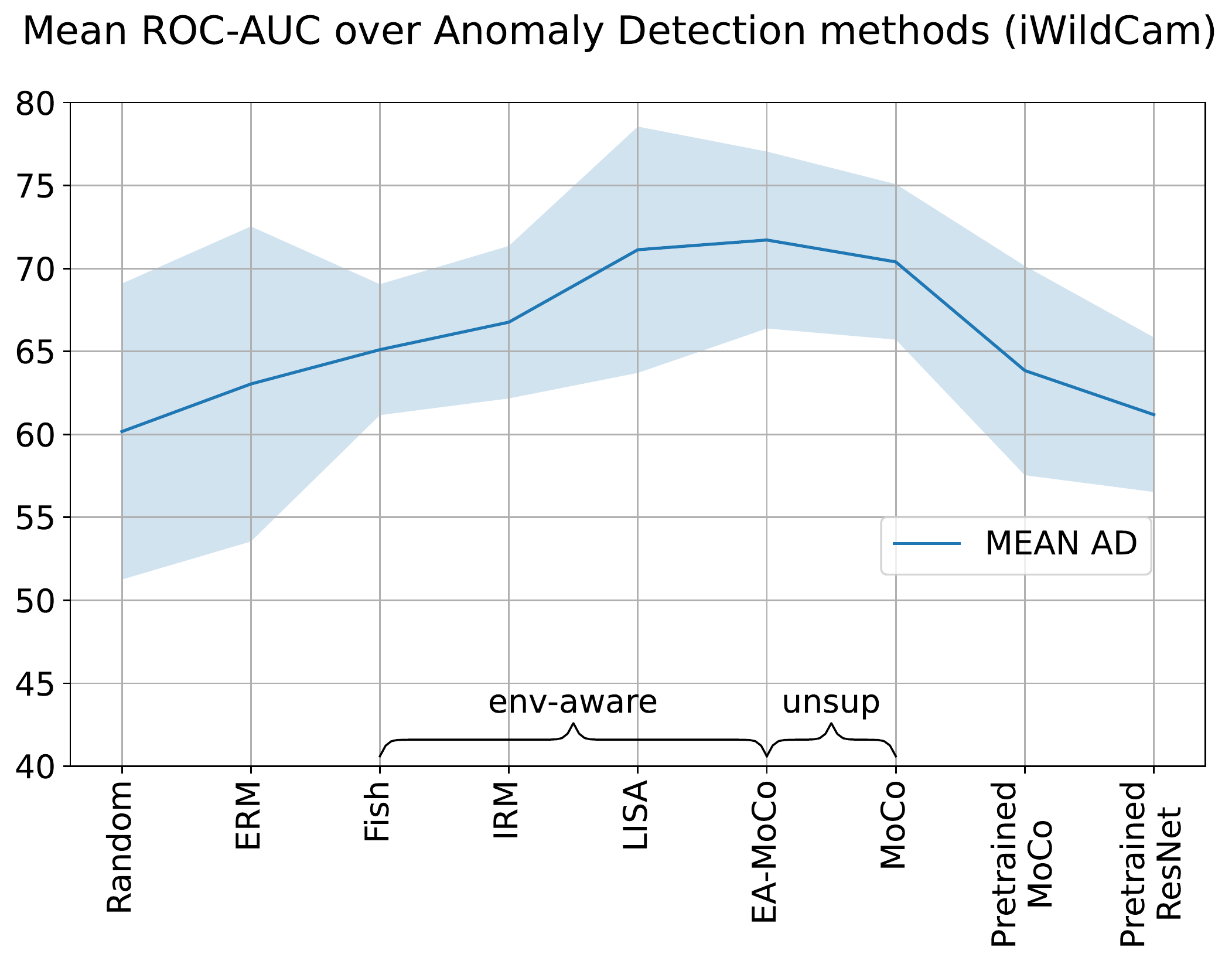}
\end{center}
\caption{Scores for multiple unsupervised AD methods (left, in different colors) and their mean (right) trained on top of embeddings obtained under different pretrainings (OX axis). Notice that env-aware ones (Fish, IRM, and LISA) show significant improvement over ERM. More, our contrastive adaptation
\textbf{EA-MoCo} for env-awareness outperforms all the others.
}
\label{fig:cmp_pretraining}
\end{figure}

\begingroup
    \setlength{\tabcolsep}{2pt} 

    \begin{table}[t]

\begin{center}
\caption{ROC-AUC scores for AD in Style OOD setup, where an anomaly is defined as Content OOD. In rows we have AD methods, applied on top of embeddings learned using pretraining algorithm named in each column. See how our env-aware contrastive solution, EA-MoCo, outperforms the others. Max values per row in bold.}
\begin{tabular}{ll c ccc ccccc}

\toprule

& \parbox[t]{2mm}{\multirow{2}{*}{\bf Pretrain}} & \multicolumn{1}{c}{\bf None} & \multicolumn{4}{c}{\bf Supervised} & \multicolumn{2}{c}{\bf Unsupervised} & \multicolumn{2}{c}{\bf Other dataset}\\

\cmidrule(lr){3-3}
\cmidrule(lr){4-7}
\cmidrule(lr){8-9}
\cmidrule(lr){10-11}
&   & \shortstack{Random} &  \shortstack{ERM} & \shortstack{Fish} & \shortstack{IRM} & \shortstack{Lisa} & \shortstack{\bf EA-MoCo} & \shortstack{MoCo v3} & \shortstack{MoCo v3} & \shortstack{ResNet} \\ 
\toprule

\parbox[t]{2mm}{\multirow{8}{*}{\rotatebox[origin=c]{90}{Anom. Detect. method}}}
& IsoForest~\cite{isoforest} & 65.2 & 63.1 & 68.0 & 64.3 & \bf 75.2 & 70.9 & 68.4 & 64.6 & 61.8 \\
 & INNE~\cite{inne} & 50.1 & 67.7 & 66.1 & 68.7 & 76.5 & \bf 77.0 & 71.9 & 68.7 & 57.8 \\
 & LODA~\cite{loda} & 65.1 & 63.8 & 66.7 & 66.2 & \bf 73.9 & 71.1 & 66.9 & 67.1 & 69.9 \\
 \cmidrule(lr){2-11}
 & OCSVM~\cite{ocsvm} & 57.9 & 67.5 & 65.5 & 64.5 & \bf 78.4 & 71.4 & 68.5 & 57.1 & 62.1 \\
 & PCA~\cite{pca} & 64.1 & 40.4 & 63.3 & 64.4 & 55.6 & \bf 67.7 & 63.9 & 60.9 & 63.2 \\
 \cmidrule(lr){2-11}
 & LOF5~\cite{lof} & 43.2 & 61.0 & 59.7 & 61.3 & 65.1 & 60.9 & \bf 68.3 & 58.5 & 53.2 \\
 & KNN~\cite{knn} & 73.2 & 75.7 & 72.0 & 77.7 & 66.9 & 77.0 & \bf 78.9 & 76.5 & 57.8 \\
 \cmidrule(lr){2-11}
 & KDE~\cite{kde} & 62.6 & 65.1 & 59.4 & 67.0 & 77.4 & \bf 77.8 & 76.3 & 57.4 & 63.6 \\
\cmidrule(lr){1-2}
 & \bf Mean AD (OOD) & 60.2 & 63.0 & 65.1 & 66.8 & 71.1 & \bf 71.7 & 70.4 & 63.8 & 61.2 \\
\bottomrule

\end{tabular}
\label{tab:main_cmp}
\end{center}
\end{table}
\endgroup


\textbf{iWildCam dataset } Ideally, for testing the robustness of an algorithm in our setup, we would need a dataset with multiple environments annotations, but also with anomalies. Since there is none tackling both aspects, we chose to start with a multi-environment dataset, on top of which we define the Content anomalies in a standard way, as classes unseen at training time \cite{ood-survey,ad_survey}. iWildCam~\cite{iwildcam} is a dataset used for studying robustness at distribution shifts, containing images from various camera traps placed in the wild, exhibiting drastic distribution shifts in image Style (\eg illumination, background, vegetation). We present in Appendix~\ref{apx:iwildcam} the details for building the AD setup on top of the dataset.

\subsection{Environment-aware algorithms in pretraining } We learn image embeddings in several ways, comparing the downstream performance of the AD methods applied in an unsupervised manner on top of them. As backbone, we use ResNet-18~\cite{resnet}. \textbf{a) Empirical Risk Minimization} (ERM~\cite{erm}) is the main baseline, which we compare against several env-aware methods: \textbf{b) Fish}~\cite{fish}, \textbf{c) Invariant Risk Minimization} (IRM~\cite{irm}), and \textbf{d) LISA}~\cite{lisa}. As shown in Tab.~\ref{tab:main_cmp}, LISA proves to be the best out of the supervised algos. Also see that in general, all embeddings learned from env-aware algorithms perform better in the downstream task: content OOD detection (anomaly). We also pretrained the embeddings using a contrastive, unsupervised solution (\textbf{e) MoCo v3}~\cite{mocov3}), which proved to be a strong baseline. Since we noticed before that env-aware capabilities improve the robustness, we add them on top of the contrastive method, as detailed in Sec.~\ref{sec:our_approach}. On average, our \textbf{Env-Aware MoCo} perform best, as seen in Fig.~\ref{fig:cmp_pretraining} and Tab.~\ref{tab:main_cmp}.


\subsection{AD methods }
For detecting anomalies, we feed the embeddings to a variety of anomaly detectors, covering most of the AD types. As \textbf{a) ensemble-based} methods, we test IsolationForest~\cite{isoforest}, INNE~\cite{inne} and LODA~\cite{loda}. For \textbf{b) linear models}, we test the classics: OCSVM~\cite{ocsvm} and PCA~\cite{pca}. The delegates for \textbf{c) proximity-based} are LOF~\cite{lof} and KNN~\cite{knn}, while for \textbf{d) probabilistic} detectors we have KDE~\cite{kde}. We use pyod~\cite{pyod} implementations and validated the dataset-related hyper-parameters (\eg number of trees, neighbours, bins, gamma, standardization). We detail in Appendix~\ref{apx:ad_params} and will make the code publicly available.






\section{Conclusions}
This work tackles the unsupervised anomaly detection task in the Style distribution shift scenario. First, we formally analyze the task in connection to existing frameworks. Next, we prove that employing env-aware pretraining methods can boost the performance of shallow AD methods in this setup. Finally, we propose an env-aware contrastive method, with up to 8.7\% improvement on iWildsCam AD setup, over the ERM baseline.

\subsubsection*{Acknowledgments}
We thank Razvan Pascanu for guiding us on how to better formalize the problem.

\bibliographystyle{plainnat}
\bibliography{biblio}

\newpage
\clearpage
\appendix

\section{Appendix}

\subsection{AD methods hyper-parameters}
\label{apx:ad_params}
We chose for each method (loaded from pyOD library~\cite{pyod}) the proper hyper-parameters for iWildCam, by maximizing the score for the basic (ERM) setup. Those are the following:
\begin{verbatim*}
"IsoForest":iforest.IForest(behaviour="new") without_scaler
"INNE":inne.INNE(n_estimators=51) without_scaler
"LODA":loda.LODA(n_bins=25, n_random_cuts=100) with_scaler
"OCSVM":ocsvm.OneClassSVM(gamma="auto") with_scaler
"PCA":pca.PCA(standardization=False, whiten=True) with_scaler
"LOF5":neighbors.LocalOutlierFactor(n_neighbors=5, novelty=True, metric='euclidean', n_jobs=-1) with_scaler
"KNN":knn.KNN(n_jobs=-1) without_scaler
"KDE":kde.KDE() without_scaler
\end{verbatim*}

\subsection{Set-up formalization details}
\label{apx:formalize-sup-unsup}
We showcase the different set-ups in which machine learning algorithms are used at present, with respect to the different types of changes these algorithms are designed to capture. In addition, we proposed a new set-up that, to our knowledge, has never been addressed before, all to be observed in Tab.~\ref{tab:formalize-sup-unsup}.

\begingroup
    \setlength{\tabcolsep}{6pt} 
\begin{table}[t]

    \begin{center}
     \caption{Shifting paradigm relation to Anomaly detection. We emphasize the current algorithms working in each paradigm. $p_e(x)$ is the probability distribution for env $e$.}
        \begin{tabular}{lccl}
        \toprule
        & \bf Style & \bf Content & \bf Description \\ 
        \toprule
        \multirow{5}{*}{\bf A.} & \multirow{5}{*}{\bf ID} & \multirow{5}{*}{\bf ID} & \textbf{Assumption:}\\
        & & & \tabindent $p_e(x_S, x_C, y)$, $p_e(x_S, x_C)$ are constant\\
        \cmidrule(lr){4-4}
         & & & \textbf{Goal/Task:}\\
         & & & \tabindent model $p_e(y|x)$ or $p_e(x, y)$ or $p_e(x)$\\
        \cmidrule(lr){4-4}
         & & & \textbf{\eg} algorithms following the ERM paradigm\\
        
        \midrule
        \multirow{5}{*}{\bf B.} & \multirow{5}{*}{\bf OOD} & \multirow{5}{*}{\bf ID} & \textbf{Assumption:} \\
        & & & \tabindent $p_e(x_S)$ changes over envs - closer to real-world scenarios \\
        \cmidrule(lr){4-4}
        & & & \textbf{Goal/Task:}\\
        & & & \tabindent same as \textbf{A.}, while being robust to Style changes \\ 
        \cmidrule(lr){4-4}
        & & & \textbf{\eg} IRM, V-Rex, Fish, Lisa\\ 
        
        \midrule
        \multirow{5}{*}{\bf C.} & \multirow{5}{*}{\bf ID} & \multirow{5}{*}{\bf OOD} & \textbf{Assumption:} \\
        & & & \tabindent $p_e(x_C)$ changes over envs\\
        \cmidrule(lr){4-4}
        & & & \textbf{Goal/Task:} \\
        & & & \tabindent detect Content changes\\
        \cmidrule(lr){4-4}
        & & & \textbf{\eg} open set recognition; detect semantic anomalies or novelties \\
        
        \midrule
        \multirow{5}{*}{\bf D.} & \multirow{5}{*}{\bf OOD} & \multirow{5}{*}{\bf OOD} & \textbf{Assumption:} \\
        & & & \tabindent both $p_e(x_S)$, $p_e(x_C)$ change over envs - closer to real-world scenarios \\
        \cmidrule(lr){4-4}
        & & & \textbf{Goal/Task:}\\
        & & & \tabindent same as \textbf{C.},while being robust to Style changes\\
        \cmidrule(lr){4-4}
        & & & \textbf{\eg} \textbf{EA-MoCo} (our approach)\\
        \bottomrule
        \end{tabular}
        \label{tab:formalize-sup-unsup}
    \end{center}
\end{table}
\endgroup

\subsection{Env-Aware MoCo algorithm: EA-MoCo}
\label{apx:ea-moco-env}
We summarize our proposed env-aware method of anomaly detection in distribution shift scenarios in an easy-to-follow algorithm. We use only one positive sample per anchor, chosen from a different, random environment, at the smallest distance defined by the previously trained autoencoder (over all envs), as described in Alg.~\ref{alg:ea-moco-env}. We do not further augment the positive sample. 

\begin{algorithm}[t]
    \caption{- \textbf{Env-aware contrastive learning}}
    \label{alg:ea-moco-env}
        $X_{e_a}$  - input samples from env $a$ \;\;\;\;\; $x_{i\_e_a}$  - input sample $i$, from env $a$\\
        $e_1, e_2, ...e_t$ - train envs \;\;\;\;\;\;\;\;\;\;\;\;\;\;\;\;\;\, $e_{t+1}, e_{t+2}, ...e_{t+k}$ - test envs
    \algrule
    \textbf{Results:} 1) style-robust embeddings $X^{SR}$; 2) anomalies prediction $Y^{AD}$
    \begin{algorithmic}[1]
        \algrule
        \Statex \textcolor{algo_comments}{// Step 1. Compute distances between all samples}
        \State $X_{e_1..e_t}^{AE} \leftarrow train\_autoencoder(X_{e_1..e_t}$) \textcolor{algo_comments}{ \;\;\;\;\;\;\;\;\;\;\;\;\;  // train an autoencoder over all training envs}
        \State $dist_{i\_e_a,j\_e_b} \leftarrow l2(x_{i\_e_a}^{AE}, x_{j\_e_b}^{AE}),  \forall x_{i\_e_a} \neq x_{j\_e_b}, a, b \in \overline{1,t}$ \;\;\;\;\; \textcolor{algo_comments}{ // use embeddings for distances}
        \Statex
        \Statex \textcolor{algo_comments}{// Step 2. Contrastive approach based on envs - Moco v3}
        \State $x_{+} \leftarrow \argmin_{\forall x_{j\_e_b} \in X_{e_b}, e_a \neq e_b} dist(x_{i\_e_a},x_{j\_e_b})$ \;\;\;\;\;\;\;\;\;\;\;\;\;\;\;\;\;\; \textcolor{algo_comments}{// closest, from another domain} 
        \State $x_{-} \leftarrow \{x_{j\_e_a} | j \sim batch_i \}$ \;\;\;\;\;\;\;\;\;\;\;\;\;\;\;\;\;\;\;\;\;\;\;\;\;\;\;\;\;\;\;\;\textcolor{algo_comments}{// usual negative samples, the rest of the batch} 
        \State $x_{i\_e_a}^{SR} \leftarrow train\_contrastive(x_{i\_e_a}, x_{-}, x_{+})$\;\;\;\;\;\;\;\;\textcolor{algo_comments}{//SR = Style-Robust; train similar to MoCov3} 
        \Statex 
        \Statex \textcolor{algo_comments}{// Step 3. Downstream task (anomaly detection)}
        \State $\omega^* \leftarrow AD(X^{SR}_{e_1...e_t})$ ;\;\;\;\;\;\;\;\;\;\;\;\;\;\;\;\;\;\;\;\;\;\;\;\;\;\;\;\;\;\;\;\;\;\; \textcolor{algo_comments}{// Train AD on the new Style-Robust embeddings } 
        \State $Y^{AD}_{e_{t+1}..e_{t+k}} \leftarrow \omega^*(X^{SR}_{e_{t+1}..e_{t+k}})$
        \;\;\;\;\;\;\;\;\;\;\;\;\;\;\;\;\;\;\;\;\;\;\;\;\;\;\;\;\;\;\;\;\;\;\;\;\;\;\;\;\;\;\;\;\;\;\;\;\;\;\;\; \textcolor{algo_comments}{// Apply AD on the testset} 
    \end{algorithmic}
\end{algorithm}

\subsection{iWildCam}
\label{apx:iwildcam}
For building the anomaly detection setup from the existing classification setup in iWildCam, we group the classes in 3 buckets (two for normality - with class label < 125 - and one with the rest, being used as anomalies). We keep only the domains present in each bucket, with sufficient samples each. Samples belonging to the normal classes are used in pretraining and AD training. The test set contains both normal and abnormal samples,  but only from envs unseen at training time (out-of-distribution from the Style point of view). We will make the split publicly available.

\end{document}

%% file: math_commands.tex
\usepackage{amsmath,amsfonts,bm}

\def\eqref#1{equation~\ref{#1}}

\def\1{\bm{1}}

\DeclareMathAlphabet{\mathsfit}{\encodingdefault}{\sfdefault}{m}{sl}
\SetMathAlphabet{\mathsfit}{bold}{\encodingdefault}{\sfdefault}{bx}{n}

\DeclareMathOperator*{\argmin}{arg\,min}